\documentclass{article}

\PassOptionsToPackage{numbers,compress,sort}{natbib}
\usepackage[final]{neurips_2019}

\usepackage[colorinlistoftodos]{todonotes}

\usepackage[utf8]{inputenc} % allow utf-8 input
\usepackage[T1]{fontenc}    % use 8-bit T1 fonts
\usepackage{hyperref}       % hyperlinks
\usepackage{url}            % simple URL typesetting
\usepackage{booktabs}       % professional-quality tables
\usepackage{amsfonts}       % blackboard math symbols
\usepackage{nicefrac}       % compact symbols for 1/2, etc.
\usepackage{microtype}      % microtypography
\usepackage{verbatim}
\usepackage{times}
\usepackage{epsfig}
\usepackage{graphicx}
\usepackage{amsmath}
\usepackage{amssymb}
\usepackage{subfigure}
\usepackage{url}
\usepackage{calc}
\usepackage{color}
\usepackage{fullpage}
\usepackage{times}
\usepackage{fancyhdr,graphicx,amsmath,amssymb}
\usepackage{tabu}
\usepackage{xcolor}
\usepackage[ruled,vlined]{algorithm2e}
\usepackage{appendix}
\usepackage{multirow}
\usepackage{mathtools}
\usepackage[nameinlink]{cleveref}
\usepackage[procnames]{listings}
\usepackage{wrapfig}

\newcommand{\methodfull}{Positional Normalization}
\newcommand{\method}{PONO}
\newcommand{\methodms}{MS}
\newcommand{\methodmsfull}{Moment Shortcut}
\newcommand{\dynamicmethodms}{DMS}
\newcommand{\dynamicmethodmsfull}{Dynamic Moment Shortcut}
\newcommand{\methodall}{PONO-MS}
\newcommand{\dynamicmethodall}{PONO-DMS}

\definecolor{bleudefrance}{rgb}{0.0, 0.36, 1.0} %;0.0, 0.44, 1.0


\title{\methodfull{}}

\author{%
  Boyi Li$^{1,2}$\thanks{: Equal contribution.}, \hspace{0.1cm} Felix Wu$^{1}$\footnotemark[1],  \hspace{0.1cm}
  Kilian Q. Weinberger$^{1}$, \hspace{0.1cm} Serge Belongie$^{1,2}$\\
  $^{1}$Cornell University $^{2}$Cornell Tech\\
  \texttt{\{bl728, fw245, kilian, sjb344\}@cornell.edu} \\
}

\begin{document}

\maketitle

\begin{abstract}
A popular method to reduce the training time of deep neural networks is to normalize activations at each layer. Although various normalization schemes have been proposed, they all follow a common theme: normalize across spatial dimensions and discard the extracted statistics. In this paper, we propose an alternative normalization method that noticeably departs from this convention and normalizes exclusively across channels. We argue that the channel dimension is naturally appealing as it allows us to extract the first and second moments of features extracted at a particular image position. These moments capture structural information about the input image and extracted features, which  opens a new avenue along which a network can benefit from feature normalization: Instead of disregarding the normalization constants, we propose to re-inject them into later layers to preserve or transfer structural information in generative networks. Codes are available at \href{https://github.com/Boyiliee/PONO}{https://github.com/Boyiliee/PONO}.

\end{abstract}

\section{Introduction}
\label{intro}

\begin{wrapfigure}{R}{0.4\linewidth}
\vspace{-3ex}
    \centering
    \includegraphics[width=\linewidth]{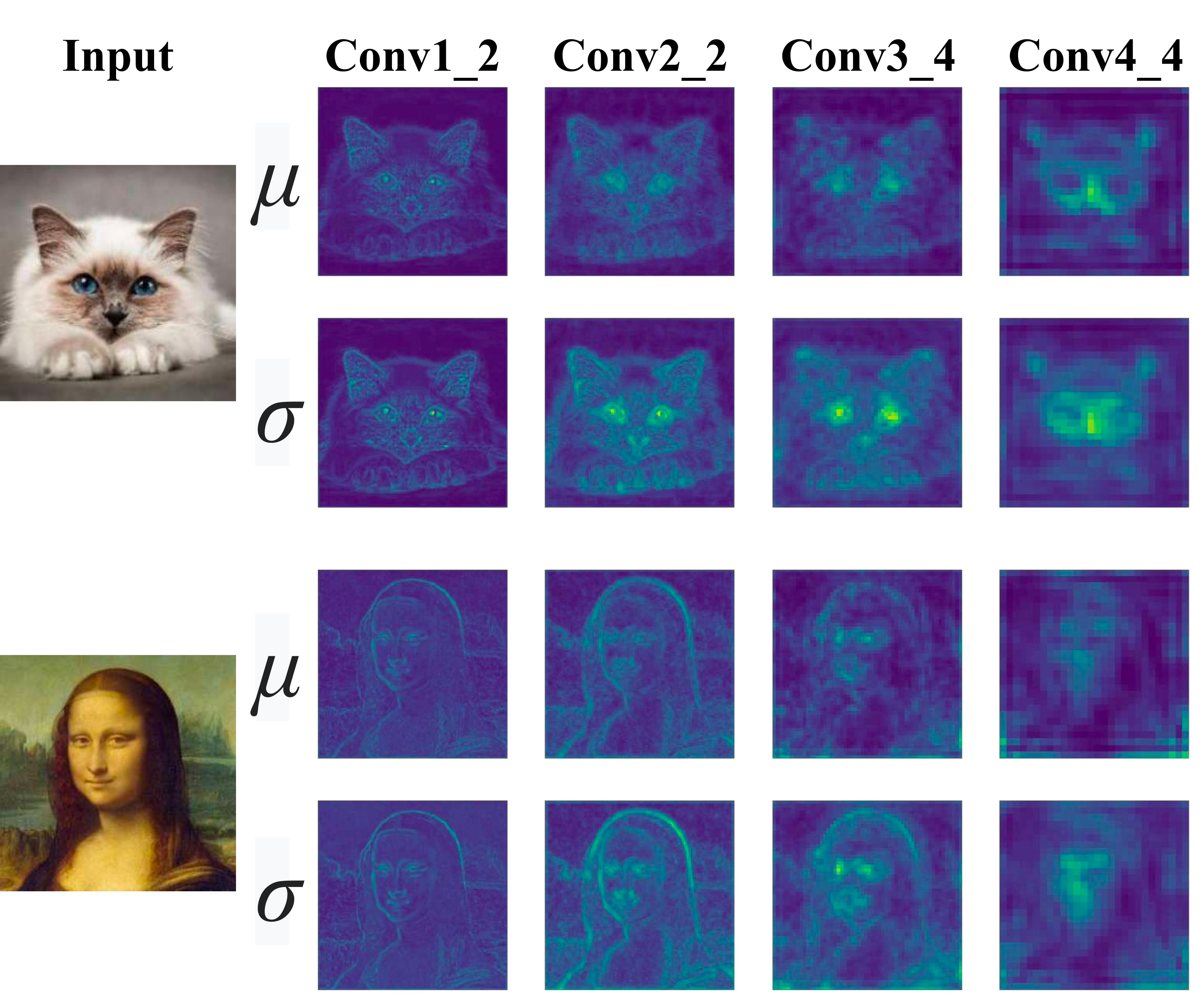}
    \caption{The mean $\mu$ and standard deviation $\sigma$ extracted by \method{} at different layers of VGG-19 capture structural information from the input images.}
    \label{fig:vgg_stats}
\vspace{-2ex}
\end{wrapfigure}

A key innovation that enabled the undeniable success of deep learning is the internal normalization of activations. Although normalizing inputs had always been one of the ``tricks of the trade'' for training neural networks~\citep{lecun2012efficient},  batch normalization (BN)~\citep{ioffe2015batch} extended this practice to every layer, which turned out to have crucial benefits for deep networks.  
While the success of normalization methods was initially attributed to ``reducing internal covariate shift'' in hidden layers~\citep{ioffe2015batch,lei2016layer}, an array of recent studies~\citep{balduzzi2017shattered,van2017l2,santurkar2018does,bjorck2018understanding,zhang2018residual,hoffer2018norm,luo2018towards,arora2018theoretical} has provided evidence that BN changes the loss surface and prevents divergence even with large step sizes~\citep{bjorck2018understanding}, which accelerates training~\citep{ioffe2015batch}. 

Multiple normalization schemes have been proposed, each with its own set of advantages: 
Batch normalization~\citep{ioffe2015batch} benefits training of deep networks primarily in computer vision tasks. Group normalization~\citep{wu2018group} is often the first choice for small mini-batch settings such as object detection and instance segmentation tasks.
Layer Normalization~\citep{lei2016layer} is well suited to sequence models, common in natural language processing.
Instance normalization~\citep{ulyanov2016instance} is widely used in image synthesis owing to its apparent ability to remove style information from the inputs.
However, all aforementioned normalization schemes follow a common theme: they normalize across spatial dimensions and discard the extracted statistics.
The philosophy behind their design is that the first two moments are considered expendable and should be removed.

In this paper, we introduce \methodfull{} (\method{}), which normalizes the activations at each position independently across the channels. 
The extracted mean and standard deviation capture the coarse structural information of an input image (see~\autoref{fig:vgg_stats}). Although removing the first two moments does benefit training, it also eliminates important information about the image, which --- in the case of a generative model --- would have to be painfully relearned in the decoder.  Instead, we propose to bypass and inject the two moments into a later layer of the network, which we refer to as \methodmsfull{} (\methodms{}) connection.

\method{} is complementary to previously proposed normalization methods (such as BN) and as such can and should be applied jointly.
We provide evidence that \method{} has the potential to substantially enhance the performance of generative models and can  exhibit favorable stability throughout the training procedure in comparison with other methods.
\method{} is designed to deal with spatial information, primarily targeted at generative~\citep{goodfellow2014generative,isola2017image} and sequential models~\citep{sutskever2014sequence,karpathy2014large,hochreiter1997long,rumelhart1986learning}. 
We explore the benefits of PONO with MS in several initial experiments across different model architectures and image generation tasks and provide code online at 
\href{https://github.com/Boyiliee/PONO}{https://github.com/Boyiliee/PONO}.

\section{Related Work}
\label{sec:related_work}

\begin{figure*}[t]
    \centering
    \includegraphics[width=\textwidth]{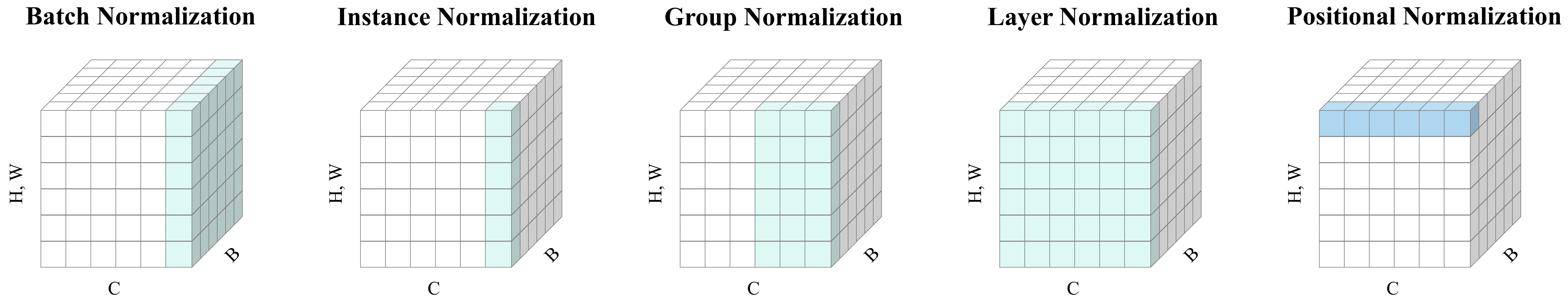}
    \caption{\methodfull{} together with previous normalization methods. In the figure, each subplot shows a feature map tensor, with $B$ as the batch axis, $C$ as the channel axis, and $(H,W)$ as the spatial axis. The entries colored in \textbf{\textcolor[HTML]{6EAC9C}{green}} or \textbf{\textcolor[HTML]{3498DB}{blue}} (ours) are normalized by the same mean and standard deviation. Unlike previous methods, our method processes each position independently, and compute both statistics across the channels.}
    \label{fig:pn_struct}
    \vspace{-0.10in}
\end{figure*}

Normalization is generally applied to improve convergence speed during training~\citep{orr2003neural}.
Normalization methods for neural networks can be roughly categorized into two regimes: \textit{normalization of weights}~\citep{NIPS2016_6114,miyato2018spectral,wu2018pay,weightstandardization} and \textit{normalization of activations}~\citep{ioffe2015batch,lei2016layer,wu2018group,lyu2008nonlinear,jarrett2009best,krizhevsky2012imagenet,ulyanov2016instance,luo2018differentiable,shao2019ssn}. 
In this work, we focus on the latter.

Given the activations $X \in \mathbb{R}^{B \times C \times H \times W}$ (where $B$ denotes the batch size, $C$ the number of channels, $H$ the height, and $W$ the width) in a given layer of a neural net, the normalization methods differ in the dimensions over which they compute the mean and variance, see~\autoref{fig:pn_struct}.
In general, activation normalization methods compute the mean $\mu$ and standard deviation (std) $\sigma$ of the features in their own manner, normalize the features with these statistics, and optionally apply an affine transformation with parameters $\beta$ (new mean) and $\gamma$ (new std). This can be written as 
\begin{equation}
X'_{b, c, h, w} = \gamma \left( \frac{X_{b, c, h, w} - \mu}{\sigma} \right) + \beta.
\end{equation}

Batch Normalization (BN)~\citep{ioffe2015batch} computes $\mu$ and $\sigma$ across the B, H, and W dimensions. BN increases the robustness of the network with respect to high learning rates and  weight initializations~\citep{bjorck2018understanding}, which in turn drastically improves the convergence rate.  
Synchronized Batch Normalization treats features of mini-batches across multiple GPUs like a single mini-batch. Instance Normalization (IN)~\citep{ulyanov2016instance} treats each instance in a mini-batch independently and computes the statistics across only spatial dimensions (H and W).
IN aims to make a small change in the stylization architecture results in a significant qualitative improvement in the generated images. 
Layer Normalization (LN) normalizes all features of an instance within a layer jointly, i.e., calculating the statistics over the C, H, and W dimensions.
LN is beneficial in natural language processing applications~\citep{lei2016layer,vaswani2017attention}.
Notably, none of the aforementioned methods normalize the information at different spatial position independently. This limitation gives rise to our proposed \methodfull{}.

Batch Normalization introduces two learned parameters $\beta$ and $\gamma$ to allow the model to adjust the mean and std of the post-normalized features. Specifically, $\beta, \gamma \in \mathbb{R}^{C}$ are channel-wise parameters. 
Conditional instance normalization (CIN)~\citep{dumoulin2017learned} keeps a set parameter of pairs $\{(\beta_i, \gamma_i) | i \in \{1, \dots, N\}  \}$ which enables the model to have $N$ different behaviors conditioned on a style class label $i$.
Adaptive instance normalization (AdaIN)~\citep{huang2017arbitrary} generalizes this to an infinite number of styles by using the $\mu$ and $\sigma$ of IN borrowed from another image as the $\beta$ and $\gamma$.
Dynamic Layer Normalization (DLN)~\citep{kim2017dynamic} relies on a neural network to generate the $\beta$ and $\gamma$. 
Later works~\citep{huang2018multimodal,karras2018style} refine AdaIN and generate the $\beta$ and $\gamma$ of AdaIN dynamically using a dedicated neural network.
Conditional batch normalization (CBN)~\citep{de2017modulating} follows a similar spirit and uses a neural network that takes text as input to predict the residual of $\beta$ and $\gamma$, which is shown to be  beneficial to visual question answering models.

Notably, all aforementioned methods generate $\beta$ and $\gamma$ as vectors, shared across spatial positions.
In contrast, Spatially Adaptive Denormalization (SPADE)~\citep{park2019SPADE},  an extension of Synchronized Batch Normalization with dynamically predicted weights, generates the spatially dependent $\beta, \gamma \in \mathbb{R}^{B \times C \times H \times W}$ using a two-layer ConvNet with raw images as inputs.

Finally, we introduce shortcut connections to transfer the first and second moment from early to later layers. Similar skip connections (with add, concat operations) have been introduced in ResNets~\citep{he2016deep} and DenseNets~\citep{huang2017densely} and earlier works~\citep{bishop1995neural,hochreiter1997long,ripley2007pattern,srivastava2015highway,kim2016accurate}, and are highly effective at improving network optimization and convergence properties~\citep{vislosslandscape18}.

\section{\methodfull{} and \methodmsfull{}}
\label{pn}

\begin{wrapfigure}{R}{0.3\linewidth}
    \vspace{-0.2in}
    \centering
    \includegraphics[width=0.9\linewidth]{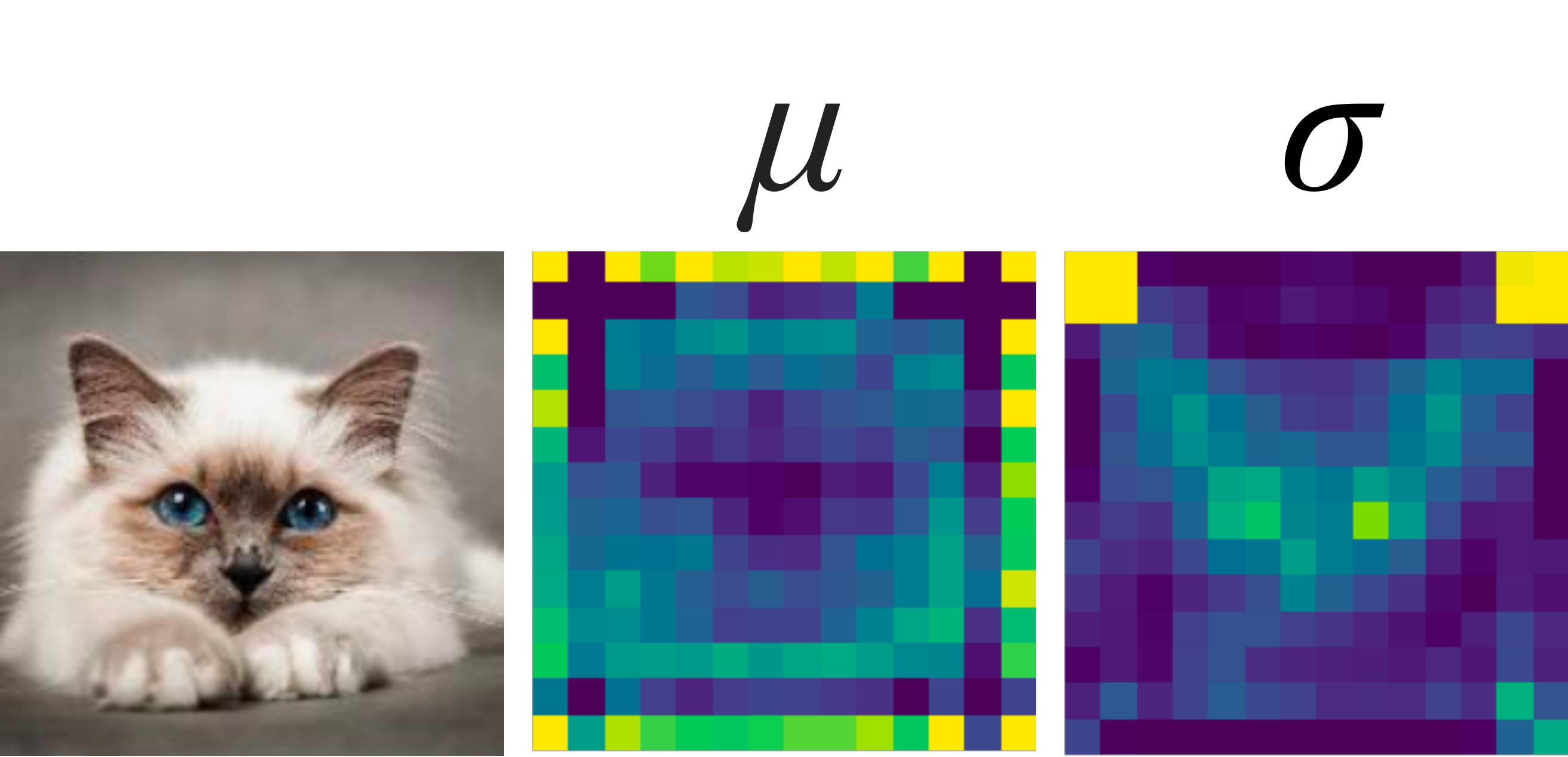}
    \caption{\method{} statistics of DenseBlock-3 of a pretrained DenseNet-161.}
    \label{fig:densenet_stats}
    %\vspace{-0.5in}
\end{wrapfigure}

Prior work has shown that feature normalization has a strong beneficial effect on the convergence behavior of neural networks~\citep{bjorck2018understanding}. Although we agree with these findings, in this paper we claim that removing the first and second order information at multiple stages throughout the network may also deprive the deep net of potentially useful information --- particularly in the context of generative models, where a plausible image needs to be generated. 

\paragraph{\method{}.} 
Our normalization scheme, which we refer to as  \emph{\methodfull{} (\method)}, differs from prior work in that we normalize exclusively over the channels at any given fixed pixel location (see \autoref{fig:pn_struct}). Consequently, the extracted statistics are  position dependent and reveal structural information at this particular layer of the deep net. The mean $\mu$ can be considered itself an ``image'', where the intensity of pixel $i,j$ represents the average activation at this particular image location in this layer. The standard deviation $\sigma$ is the natural second order extension.  
Formally, \method{} computes
\begin{equation}
\mu_{b, h, w} = \frac{1}{C}\sum_{c=1}^C X_{b, c, h, w},\ \ \sigma_{b, h, w} = \sqrt{ \frac{1}{C}\sum_{c=1}^C \left(X_{b, c, h, w} - \mu_{b, h, w}\right)^2 + \epsilon},
\end{equation}
where $\epsilon$ is a small stability constant (\emph{e.g.}, $\epsilon=10^{-5}$) to avoid divisions by zero and imaginary values due to numerical inaccuracies. 
%and is explicitly intended to capture local information about each pixel location.

%We start with the premise that the normalization constant should not be considered disposable. 
%Instead, we explicitly propose to normalize the activations within a layer exclusively across the channel direction in order to capture meaningful statistics about the extracted features. 

\paragraph{Properties.} As \method{} computes the normalization statistics at all spatial positions  independently from each other (unlike BN, LN, CN, and GN)  it is translation, scaling, and rotation invariant. Further, it is complementary to existing normalization methods and, as such, can be readily applied in combination with e.g. BN.

\paragraph{Visualization.} As the extracted mean and standard deviations are themselves images, we can visualize them to obtain information about the extract features at the various layers of a convolutional network. Such visualizations can be revealing and could potentially be used to debug or improve network architectures. 
\autoref{fig:vgg_stats} shows heat-maps of the $\mu$ and $\sigma$ captured by \method{} at several layers (Conv1\_2, Conv2\_2, Conv3\_4, and Conv4\_4) of VGG-19~\citep{simonyan2014very}. 
The figure reveals that the features in lower layers capture the silhouette of a cat while higher layers locate the position of noses, eyes, and the end points of ears —- suggesting that later layers may focus on higher level concepts corresponding to essential facial features (eyes, nose, mouth), whereas earlier layers predominantly extract generic low level features like edges. We also observe a similar phenomenon from the features of ResNets~\citep{he2016deep} and DenseNets~\citep{huang2017densely} (see \autoref{fig:densenet_stats} and Appendix).
The resulting images are reminiscent of related statistics captured in texture synthesis~\citep{freeman1991design,osada2002shape,dryden2014shape,efros1999texture,efros2001image,heeger1995pyramid,wei2000fast}. 
We observe that unlike VGG and ResNet, DenseNet exhibits strange behavior on corners and boundaries which may degrade performance when fine-tuned on tasks requiring spatial information such as object detection or segmentation. This suggests that the padding and downsampling procedure of DenseNet should be revisited and may lead to improvements if fixed, see~\autoref{fig:densenet_stats}. \label{spn_vs_in}
The visualizations of the \method{} statistics support our hypothesis that the mean $\mu$ and the standard deviation $\sigma$ may indeed capture structural information of the image and extracted features, similar to the way statistics computed by IN have the tendency to capture aspects of the style of the input image~\citep{ulyanov2016instance,huang2017arbitrary}. This extraction of valuable information motivates  the \methodmsfull{}{} described in the subsequent section.

\subsection{\methodmsfull{}}

\begin{figure*}
    \centering
    \includegraphics[width=\linewidth]{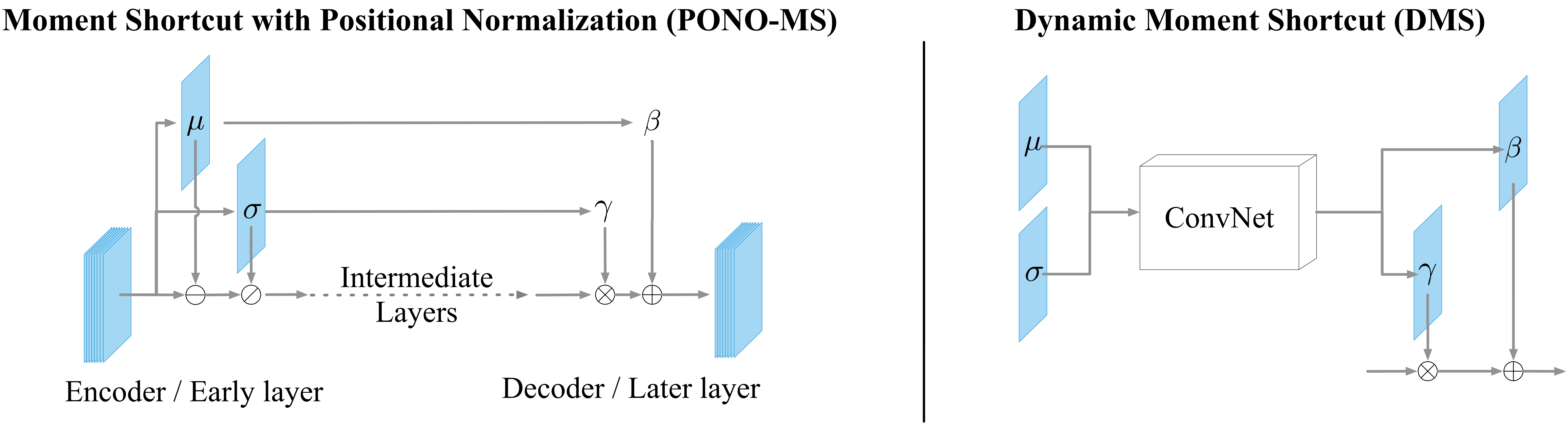}
    \caption{
    Left: \methodall{} directly uses the extracted mean and standard deviation as $\beta$ and $\gamma$.
    Right: Optionally, one may use a (shallow) ConvNet to predict $\beta$ and $\gamma$ dynamically based on $\mu$ and $\sigma$.
    }
    \label{fig:pn_arch}
    \vspace{-0.1in}
\end{figure*}

In generative models, a deep net is trained to generate an output image from some inputs (images). Typically, generative models follow an encoder-decoder architecture, where the encoder digests an image into a condensed form and the decoder recovers a plausible image with some desired properties. For example, Huang et al.~\citep{huang2017arbitrary} try to transfer the style from an image A to an image B, Zhu et al.~\citep{zhu2017unpaired} ``translate'' an image from an input distribution (e.g., images of zebras) to an output distribution (e.g., images of horses), Choi et al.~\citep{choi2018stargan} use a shared encoder-decoder with a classification loss in the encoded latent space to enable translation across multiple distributions, and \citep{huang2018multimodal,drit} combine the structural information of an image with the attributes from another image to generate a fused output.

%\paragraph{\methodmsfull{} (\methodms{}).} 
U-Nets~\citep{ronneberger2015u} famously achieve strong results and compelling optimization properties in generative models through the introduction of skip connections from the encoder to the decoder. \method{} gives rise to an interesting variant of such skip connections. Instead of connecting all channels, we only ``fast-forward'' the positional moment information $\mu$ and $\sigma$ extracted from earlier layers.  We refer to this approach as \methodmsfull{} (\methodms{}).

\paragraph{Autoencoders.} \autoref{fig:pn_arch} (left) illustrates the use of \methodms{} in the context of an autoencoder. Here, we extract the first two moments of the activations ($\mu,\sigma$) in an encoder layer, and send them to a corresponding decoder layer. Importantly, the mean is \emph{added} in the encoder, and the std is \emph{multiplied}, similar to $(\beta,\gamma)$  in the standard BN layer. 
To be specific,
$
\mathrm{\methodms{}}(\mathbf{x}) = \gamma F(\mathbf{x}) + \beta
$, where $F$ is modeled by the intermediate layers, and the $\beta$ and $\gamma$ are the $\mu$ and $\sigma$ extracted from the input $\mathbf{x}$. 
\methodms{} biases the decoder explicitly so that the activations in the decoder layers give rise to similar statistics than corresponding layers in the encoder. As \methodms{} shortcut connections can be used with and without normalization, we refer to  the combination of \method{} with \methodms{} as  \textbf{\methodall{}} throughout. 

Provided \method{} does capture essential structural signatures from the input images, we can use the extracted moments to transfer this information from a source to a target image. This opens an opportunity to go beyond autoencoders and use \methodall{} in image-to-image translation settings, for example in the context of CycleGAN~\citep{zhu2017unpaired} and Pix2Pix~\citep{isola2017image}. Here, we transfer the structure (through $\mu$ and $\sigma$) of one image from the encoder to the decoder of another image.

\paragraph{\dynamicmethodmsfull{}.} 
Inspired by Dynamic Layer Normalization and similar works~\citep{kim2017dynamic,huang2018multimodal,karras2018style,chen2018self,park2019SPADE}, we propose a natural extension called \dynamicmethodmsfull{} (\dynamicmethodms{}): instead of re-injecting  $\mu$ and $\sigma$ as is, we use a convolutional neural network that takes $\mu$ and $\sigma$ as inputs to generate the $\beta$ and $\gamma$ for \methodms{}. 
This network can either generate one-channel outputs $\beta, \gamma \in \mathcal{R}^{B\times 1 \times H \times W}$ or multi-channel outputs $\beta, \gamma \in \mathcal{R}^{B \times C \times H \times W}$ (like \citep{park2019SPADE}).
The right part of \autoref{fig:pn_arch} illustrates \dynamicmethodms{} with one-channel output. \dynamicmethodms{} is particularly helpful when the task involves shape deformation or distortion.
We refer to this approach as \textbf{\dynamicmethodall{}} in the following sections. In our experiments, we explore using a ConvNet with either one or two layers.

\section{Experiments and Analysis}
\label{sec:analysis}
We conduct our experiments on unpaired and paired image translation tasks using CycleGAN~\citep{zhu2017unpaired} and Pix2pix~\citep{isola2017image} as baselines, respectively. Our code is available at \href{https://github.com/Boyiliee/PONO}{https://github.com/Boyiliee/PONO}.

\subsection{Experimental Setup}
 We follow the same setup as CycleGAN~\citep{zhu2017unpaired} and Pix2pix~\citep{isola2017image} using their official code base.\footnote{\href{https://github.com/junyanz/pytorch-CycleGAN-and-pix2pix}{https://github.com/junyanz/pytorch-CycleGAN-and-pix2pix}}
 We use four datasets:
1) \textbf{Maps} (Maps $\leftrightarrow$ aerial photograph) including 1096 training images scraped from Google Maps and 1098 images in each domain for testing. 
2) \textbf{Horse $\leftrightarrow$ Zebra} including 1067 horse images and 1334 zebra images downloaded from ImageNet~\citep{deng2009imagenet} using keywords wild horse and zebra, and 120 horse images and 140 zebra images for testing.
3) \textbf{Cityscapes} (Semantic labels $\leftrightarrow$ photos)~\citep{Cordts2016Cityscapes} including 2975 images from the Cityscapes training set for training and 500 images in each domain for testing.
4) \textbf{Day $\leftrightarrow$ Night} including 17,823 natural scene images from Transient Attributes dataset~\citep{laffont2014transient} for training, and 2,287 images for testing.
The first, third, and fourth are paired image datasets; the second is an unpaired image dataset. We use the first and second for CycleGAN, and all the paired-image datasets for Pix2pix.

\paragraph{Evaluation metrics.} We use two evaluation metrics, as follows.
(1) Fr\'echet Inception Distance~\citep{heusel2017gans} between the output images and all test images in the target domain. FID uses an Inception~\citep{szegedy2015going} model pretrained on ImageNet~\citep{deng2009imagenet} to extract image features. Based on the means and covariance matrices of the two sets of extracted features, FID is able to estimate how different two distributions are.
(2) Average Learned Perceptual Image Patch Similarity distance~\citep{zhang2018perceptual} of all output and target image pairs. LPIPS is based on pretrained AlexNet~\citep{krizhevsky2012imagenet} features\footnote{\href{https://github.com/richzhang/PerceptualSimilarity}{https://github.com/richzhang/PerceptualSimilarity}, version 0.1.}, which has been shown~\citep{zhang2018perceptual} to be highly correlated to human judgment.

\paragraph{Baselines.} We include four baseline approaches: 
(1) CycleGAN or Pix2pix baselines;
(2) these baselines with SPADE~\citep{park2019SPADE}, which passes the input image through a 2-layer ConvNet and generates the $\beta$ and $\gamma$ for BN in the decoder.
(3) the baseline with \emph{additive skip connections} where encoder activations are added to decoder activations;
(4) the baseline with \emph{concatenated skip connections}, where encoder activations are concatenated to decoder activations as additional channels (similar to U-Nets~\citep{ronneberger2015u}).
For all models, we follow the same setup as CycleGAN~\citep{zhu2017unpaired} and Pix2pix~\citep{isola2017image} using their implementations. 
Throughout we use the hyper-parameters suggested by the original authors.

\subsection{Comparison against Baselines}

We add \methodall{} and \dynamicmethodall{} to the CycleGAN generator; see the Appendix for the model architecture.
\autoref{tab:com_methods} shows that
both cases outperform all baselines at transforming maps into photos, with the only exception of SPADE (which however performs worse in the other direction).

Although skip connections could help make up for the lost information, we postulate that directly adding the intermediate features back may introduce too much unnecessary information and might distract the model.
Unlike the skip connections, SPADE uses the input to predict the parameters for normalization.
However, on Photo~$\rightarrow$~Map, the model has to learn to compress the input photos and extract structural information from it. A re-introduction of the original raw input may disturb this process and explain the worse performance.
In contrast, \methodall{} normalizes exclusively across channels which allows us to capture structural information of a particular input image and re-inject/transfer it to later layers.

\begin{table}
  \centering
  \resizebox{\textwidth}{!}{\begin{tabular}{l|l|cc|cc}
  \toprule
   &  & \textbf{Map $\rightarrow$ Photo}& \textbf{Photo $\rightarrow$ Map}& \textbf{Horse $\rightarrow$ Zebra} & \textbf{Zebra $\rightarrow$ Horse}\\ 
       Method & \# of param. & FID&FID& FID & FID\\ 
    \midrule
    CycleGAN (Baseline)&2$\times$11.378M&57.9&58.3&86.3&155.9\\
    \midrule
    +Skip Connections& \textbf{+0M}&   83.7 & 56.0&75.9&145.5\\
    +Concatenation & +0.74M &      58.9& 61.2&85.0&145.9\\
    +SPADE& +0.456M&\textbf{48.2}& 59.8&71.2&159.9\\
    \midrule
    \color{bleudefrance}+\methodall{}&\color{bleudefrance}\textbf{+0M}&       \color{bleudefrance}52.8& \color{bleudefrance}\textbf{53.2}&\color{bleudefrance}71.2&\color{bleudefrance}142.2\\
    \color{bleudefrance}+\dynamicmethodall{}&\color{bleudefrance}+0.018M&       \color{bleudefrance}53.7& \color{bleudefrance}54.1&\color{bleudefrance}\textbf{65.7}&\color{bleudefrance}\textbf{140.6}\\
    \bottomrule
  \end{tabular}}
  \caption{FID of CycleGAN and its variants on Map $\leftrightarrow$ Photo and Zebra $\leftrightarrow$ Horse datasets. CycleGAN is trained with two directions together, it is essential to have good performance in both directions.}
  \label{tab:com_methods}
%\vspace{-0.1in}
\end{table}

The Pix2pix model~\citep{isola2017image} is a conditional adversarial network introduced as a
general-purpose solution for image-to-image translation problems. Here we conduct experiments on whether \methodall{} helps Pix2pix~\citep{isola2017image} with Maps~\citep{zhu2017unpaired}, Cityscapes~\citep{Cordts2016Cityscapes} and  Day $\leftrightarrow$ Night~\citep{laffont2014transient}. We train for 200 epochs and compare the results with/without \methodall{}, under similar conditions with matching number of parameters. Results are summarized in \autoref{tab:com_pix2pix}. 

\begin{table}
  \centering
  \resizebox{\textwidth}{!}{
  \begin{tabular}{l|cc|cc|cc}
    \toprule
     & \multicolumn{2}{c|}{\textbf{Maps}~\citep{zhu2017unpaired}} & \multicolumn{2}{c|}{\textbf{Cityscapes}~\citep{Cordts2016Cityscapes}} &  \multicolumn{2}{c}{\textbf{Day $\leftrightarrow$ Night}~\citep{laffont2014transient}}  \\
     & \textbf{Map $\rightarrow$  Photo} &\textbf{Photo $\rightarrow$ Map} & \textbf{SL $\rightarrow$ Photo} &\textbf{Photo $\rightarrow$ SL} &  \textbf{Day $\rightarrow$ Night}&\textbf{Night $\rightarrow$Day}  \\
    \midrule
    Pix2pix (Baseline) &60.07\ /\ \textbf{0.333} &68.73\ /\ 0.169&71.24\ /\ 0.422&102.38\ /\ \textbf{0.223}&196.58\ /\ 0.608&131.94\ /\ \textbf{0.531}\\
    
    \color{bleudefrance}+\methodall{} &\color{bleudefrance}\textbf{56.88}\ /\ \textbf{0.333}&\color{bleudefrance}\textbf{68.57}\ /\ \color{bleudefrance}\textbf{0.166}   &\color{bleudefrance}\textbf{60.40}\ /\ \color{bleudefrance}\textbf{0.331} &\color{bleudefrance}\textbf{97.78}\ /\ \color{bleudefrance}0.224&\color{bleudefrance}\textbf{191.10}\ /\ \color{bleudefrance}\textbf{0.588}&\color{bleudefrance}\textbf{131.83}\ /\ 0.534\\
    \bottomrule
    
  \end{tabular}
  }
  \caption{Comparison based on Pix2pix by FID\ /\ LPIPS on Maps~\citep{zhu2017unpaired}, Cityscapes~\citep{Cordts2016Cityscapes}, and Day2Night. Note: for all scores, the lower the better (SL is short for \emph{Semantic labels}).}
  \label{tab:com_pix2pix}
  \vspace{-0.2in}
\end{table}

\subsection{Ablation Study}
\autoref{tab:com_diffconfig} contains the results of several experiments to evaluate the sensitivities and  design choices of \methodall{} and \dynamicmethodall{}.
Further, we evaluate \emph{Moment Shortcut (MS)} without \method{}, where we bypass  both statistics, $\mu$ and $\sigma$, without normalizing the features. The results indicate  that \methodall{} outperforms \methodms{} alone, which suggests that normalizing activations with \method{} is beneficial.
\dynamicmethodall{} can lead to further improvements, and some settings (e.g. \emph{1 conv 3 $\times$ 3, multi-channel}) consistently outperform \methodall{}. 
Here, multi-channel predictions are clearly superior over single-channel predictions but  
we do not observe consistent improvements from a $5\times 5$ rather than a $3\times 3$ kernel size.

\begin{table}
  \centering
  \resizebox{\textwidth}{!}{\begin{tabular}{lcccc}
       Method &\textbf{Map $\rightarrow$ Photo}& \textbf{Photo $\rightarrow$ Map}& \textbf{Horse $\rightarrow$ Zebra} & \textbf{Zebra $\rightarrow$ Horse}\\ 
    \toprule
    CycleGAN (Baseline)&57.9&58.3&86.3&155.9\\
    \midrule
    +\methodmsfull{} (\methodms)&   54.5    & 56.6 &79.8&146.1\\
    \midrule
    +\methodall{} &       52.8& 53.2&71.2&142.2\\
    +\dynamicmethodall{} (1 conv $3\times3$, one-channel)&       55.1& 53.8&74.1&147.2\\
    +\dynamicmethodall{} (2 conv $3\times3$, one-channel)&      56.0 &53.3 &81.6&144.8\\
    +\dynamicmethodall{} (1 conv $3\times3$, multi-channel)&       53.7& 54.1&65.7&\textbf{140.6}\\
    +\dynamicmethodall{} (2 conv $5\times5$, multi-channel)&      52.7 & 54.7&\textbf{64.9}&155.2\\
    +\dynamicmethodall{} (2 conv $3\times3, 5\times5$, multi-channel)&     \textbf{48.9}  & 57.3&74.3&148.4\\
    +\dynamicmethodall{} (2 conv $3\times3$, multi-channel) & 50.3 & \textbf{51.4} &72.2&146.1\\
    \bottomrule
  \end{tabular}}
  \caption{Comparisons of ablation study on FID (lower is better). \methodall{} outperforms \methodms{} alone. \dynamicmethodall{} can help obtain better performance than \methodall{}. }
   \vspace{-0.1in}
   
  \label{tab:com_diffconfig}
\end{table}

\paragraph{Normalizations.}
Unlike previous normalization methods such as BN and GN that emphasize on accelerating and stabilizing the training of networks, PONO is used to split off part of the spatial information and re-inject it later. Therefore, \methodall{} can be applied jointly with other normalization methods. In \autoref{tab:com_diffnorm} we evaluate four normalization approaches (BN, IN, LN, GN) with and without \methodall{}, and \methodall{} without any additional normalization (bottom row). In detail, \textit{BN + PONO-MS} is simply applying \textit{PONO-MS} to the baseline model and keep the original BN modules which have a different purpose: to stabilize and speed up the training. We also show the models where BN is replaced by LN/IN/GN as well as these models with PONO-MS. The last row shows PONO-MS can work independently when we remove the original BN in the model.
Each table entry displays the FID score without and with \methodall{} (the lower score is in \textbf{bold}). The final column (very right) contains the average improvement across all four tasks, relative to the default architecture, BN without \methodall{}. 
Two clear trends emerge: 1. All four normalization methods improve with \methodall{} on average and on almost all individual tasks; 2. additional normalization is clearly beneficial over pure \methodall{} (bottom row). 

\begin{table}
  \centering
  \resizebox{\textwidth}{!}{
  \begin{tabular}{lllll|r}
    Method     &  \textbf{Map $\rightarrow$ Photo} & \textbf{Photo $\rightarrow$ Map} & \textbf{Horse $\rightarrow$ Zebra} & \textbf{Zebra $\rightarrow$ Horse} & Avg. Improvement\\
    \toprule
     BN (Default)\ /\ \color{bleudefrance}BN + \methodall{}&57.92\ /\ \color{bleudefrance}\textbf{52.81}&58.32\ /\ \color{bleudefrance}\textbf{53.23}&86.28\ /\ \color{bleudefrance}\textbf{71.18}&155.91\ /\ \color{bleudefrance}\textbf{142.21}
     & 1 /\ \color{bleudefrance}{\textbf{0.890}} \\
     \midrule
    IN\ /\ \color{bleudefrance}IN + \methodall{}&  67.87\ /\ \color{bleudefrance}\textbf{47.14} & 57.93\ /\ \color{bleudefrance}\textbf{54.18}&\textbf{67.85}\ /\ \color{bleudefrance}69.21&154.15\ /\ \color{bleudefrance}\textbf{153.61}
    & 0.985 /\ \color{bleudefrance}{\textbf{0.883}}\\
     \midrule
    LN\ /\ \color{bleudefrance}LN + \methodall{}&     54.84 \ /\ \color{bleudefrance}\textbf{49.81} & 53.00\ /\ \color{bleudefrance}\textbf{50.08}&87.26\ /\ \color{bleudefrance}\textbf{67.63}&154.49\ /\ \color{bleudefrance}\textbf{142.05}
    & 0.964 /\ \color{bleudefrance}{\textbf{0.853}}\\
    \midrule
    GN\ /\ \color{bleudefrance}GN + \methodall{} &51.31\ /\ \color{bleudefrance}\textbf{50.12}&50.62\ /\ \color{bleudefrance}\textbf{50.50}&93.58\ /\ \color{bleudefrance}\textbf{63.53}&\textbf{143.56}\ /\ \color{bleudefrance}144.99
    & 0.940 /\ \color{bleudefrance}{\textbf{0.849}}\\
    \midrule
    \color{bleudefrance}\methodall{} &\color{bleudefrance}\textbf{49.59}&\color{bleudefrance}\textbf{52.21}&\color{bleudefrance}\textbf{84.68}&\color{bleudefrance}\textbf{143.47} & 
   \color{bleudefrance}{\textbf{0.913}}\\
    \bottomrule
  \end{tabular}
  }
  \caption{FID scores (lower is better) of CycleGAN with different normalization methods. }.
  \label{tab:com_diffnorm}
  \vspace{-0.35in}
\end{table}

%!TEX root=../main.tex
\section{Further Analysis and Explorations}
\label{exploration}
In this section, we apply \methodall{} to two state-of-the-art unsupervised image-to-image translation models: MUNIT~\citep{huang2018multimodal} and DRIT~\citep{drit}.
Both approaches may arguably be considered concurrent works and share a similar design philosophy. Both aim to translate an image from a source to a target domain, while imposing the attributes (or the style) of another target domain image. 

As task, we are provided with an image $\mathbf{x}_A$ in source domain A and an image $\mathbf{x}_B$ in target domain B.
DRIT uses two encoders, one to extract content features $\mathbf{c}_A$ from $\mathbf{x}_A$, and the other to extract attribute features $\mathbf{a}_B$ from $\mathbf{x}_B$.
A decoder then takes $\mathbf{c}_A$ and $\mathbf{a}_B$ as inputs to generate the output image $\mathbf{x}_{A \rightarrow B}$.
MUNIT follows a similar pipeline. Both approaches are trained on the two directions, $A \rightarrow B$ and $B \rightarrow A$, simultaneously. We apply \method{} to DRIT or MUNIT immediately after the first three convolution layers (convolution layers before the residual blocks) of the content encoders. We then use \methodms{} before the last three transposed convolution layers with matching decoder sizes. We follow the DRIT and MUNIT frameworks and consider the extracted statistics ($\mu$'s and $\sigma$'s) as part of the content tensors.

\subsection{Experimental Setup}
We consider two datasets provided by the authors of DRIT:
1) \textbf{Portrait $\leftrightarrow$ Photo}~\citep{drit,liu2015deep} with 1714 painting images and 6352 human photos for training, and 100 images in each domain for testing and
2) \textbf{Cat $\leftrightarrow$ Dog}~\citep{drit} containing 771 cat images and 1264 dog images for training, and 100 images in each domain for testing.

In the following experiments, we use the official codebases\footnote{\href{https://github.com/NVlabs/MUNIT/}{https://github.com/NVlabs/MUNIT/} and \href{https://github.com/HsinYingLee/DRIT}{https://github.com/HsinYingLee/DRIT}}, closely follow their proposed hyperparameters and train all models for 200K iterations. 
We use the holdout test images as the inputs for evaluation. For each image in the source domain, we randomly sample 20 images in the target domain to extract the attributes and generate 20 output images.
We consider four evaluation metrics: 1) FID~\citep{heusel2017gans}: Fr\'echet Inception Distance between the output images and all test images in the target domain, 2) LPIPS\textsubscript{attr}~\citep{zhang2018perceptual}: average LPIPS distance between each output image and its corresponding input image in the target domain, 3) LPIPS\textsubscript{cont}: average LPIPS distance between each output image and its input in the source domain, and 4) perceptual loss (VGG)~\citep{simonyan2014very,johnson2016perceptual}: L1 distance between the VGG-19 Conv4\_4 features~\citep{chen2018cartoongan} of each output image and its corresponding input in the source domain.
The FID and LPIPS\textsubscript{attr} are used to estimate how likely the outputs are to belong to the target domain, while LPIPS\textsubscript{cont} and VGG loss are adopted to estimate how much the outputs preserve the structural information in the inputs. All of them are distance metrics where lower is better. The original implementations of DRIT and MUNIT assume differently sized input images (216x216 and 256x256, respectively), which precludes a direct comparison across approaches. 

\subsection{Results of Attribute Controlled Image Translation}
\begin{figure*}
    \includegraphics[width=\linewidth]{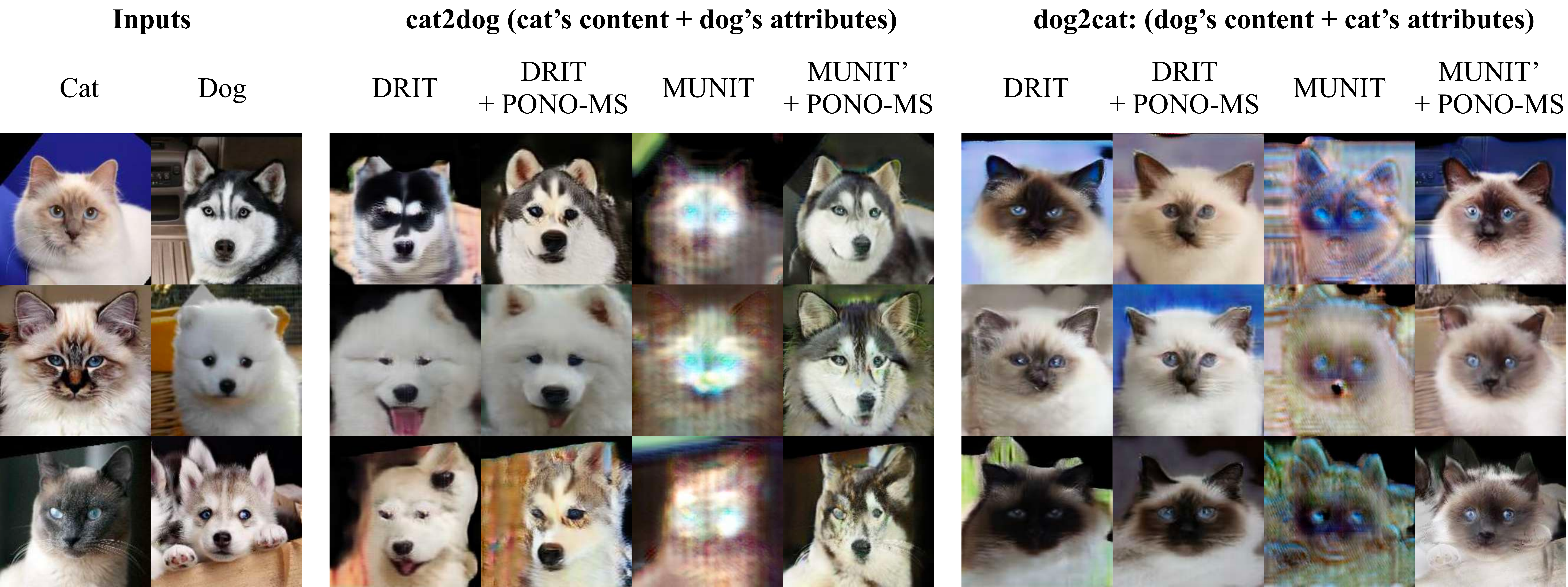}
    
    \caption{ 
    \methodall{} improves the quality of both DRIT~\citep{drit} and MUNIT~\citep{huang2018multimodal} on Cat $\leftrightarrow$ Dog.}
    \label{fig:munit_cat2dog_show}
    %\vspace{-0.20in}
\end{figure*}

\autoref{fig:munit_cat2dog_show} shows the qualitative results on the Cat $\leftrightarrow$ Dog dataset. (Here we show the results of MUNIT' + \methodall{} which will be explained later.)
We observe a clear trend that \methodall{} helps these two models obtain more plausible results. We observe the models with \methodall{} is able to capture the content features and attributes distributions, which motivates baseline models to digest different information from both domains. For example, in the first row, when translating cat to dog, DRIT with \methodall{} is able to capture the cat's facial expression, and MUNIT with \methodall{} could successfully generate dog images with plausible content, which largely boosts the performance of the baseline models. More qualitative results of randomly selected inputs are provided in the Appendix.

\autoref{tab:drit_munit} show the quantitative results on both Cat $\leftrightarrow$ Dog and Portrait $\leftrightarrow$ Photo datasets.
\methodall{} improves the performance of both models on all instance-level metrics (LPIPS\textsubscript{attr}, LPIPS\textsubscript{cont}, and VGG loss). However, the dataset-level metric, FID, doesn't improve too much. We believe the reason is that FID is calculated based on the first two order statistic of Inception features and may discard some subtle differences between each output pair.

\begin{table}
  \centering
  \resizebox{\textwidth}{!}{
  \begin{tabular}{l|rrrr|rrrr}
    \toprule
     & \multicolumn{4}{c|}{\textbf{Portrait $\rightarrow$ Photo}}  & \multicolumn{4}{c}{\textbf{Portrait $\leftarrow$ Photo}} \\
    & FID & LPIPS\textsubscript{attr} & LPIPS\textsubscript{cont} & VGG
    & FID & LPIPS\textsubscript{attr} & LPIPS\textsubscript{cont} & VGG
    \\
    \midrule
    DRIT & 131.2 & 0.545 & 0.470 & 1.796 & 104.5 & 0.585 & 0.476 & 2.033\\
    \color{bleudefrance}DRIT + \methodall{}  & \color{bleudefrance} \textbf{127.9} & \color{bleudefrance}\textbf{0.534} & \color{bleudefrance}\textbf{0.457} & \color{bleudefrance}\textbf{1.744} & \color{bleudefrance}\textbf{99.5} & \color{bleudefrance}\textbf{0.575} & \color{bleudefrance}\textbf{0.463} & \color{bleudefrance}\textbf{2.022}\\
    \midrule
    MUNIT & 220.1 & 0.605 & 0.578 & 1.888 & 149.6 & 0.619 & 0.670 & 2.599\\
    \color{bleudefrance}MUNIT + \methodall{} & \color{bleudefrance}270.5 & \color{bleudefrance}0.541 & \color{bleudefrance}0.423 & \color{bleudefrance}1.559 & \color{bleudefrance}127.5 & \color{bleudefrance}0.586 & \color{bleudefrance}0.477 & \color{bleudefrance}2.202\\
    MUNIT' & 245.0 & 0.538 & 0.455 & 1.662 & 158.1 & 0.601 & 0.620 & 2.434\\
    \color{bleudefrance}MUNIT' + \methodall{} & \color{bleudefrance} \textbf{159.4} & \color{bleudefrance}\textbf{0.424} & \color{bleudefrance}\textbf{0.319} & \color{bleudefrance}\textbf{1.324} & \color{bleudefrance}\textbf{125.1} & \color{bleudefrance}\textbf{0.566} & \color{bleudefrance}\textbf{0.312} & \color{bleudefrance}\textbf{1.824}\\
    \toprule
    \ & \multicolumn{4}{c|}{\textbf{Cat $\rightarrow$ Dog}} &  \multicolumn{4}{c}{\textbf{Cat $\leftarrow$ Dog}} \\
    & FID & LPIPS\textsubscript{attr} & LPIPS\textsubscript{cont} & VGG
    & FID & LPIPS\textsubscript{attr} & LPIPS\textsubscript{cont} & VGG
    \\
    \midrule
    DRIT & \textbf{45.8} & 0.542 & 0.581 & \textbf{2.147} & 42.0 & 0.524 & \textbf{0.576} & 2.026\\
    \color{bleudefrance}DRIT + \methodall{} & \color{bleudefrance} 47.5 & \color{bleudefrance}\textbf{0.524} & \color{bleudefrance}\textbf{0.576} & \color{bleudefrance}\textbf{2.147} & \color{bleudefrance}\textbf{41.0} & \color{bleudefrance}\textbf{0.514} & \color{bleudefrance}0.604 & \color{bleudefrance}\textbf{2.003} \\
    \midrule
    MUNIT & 315.6 & 0.686 & 0.674 & 1.952 & 290.3 & 0.629 & 0.591 & 2.110\\
     \color{bleudefrance}MUNIT + \methodall{} &  \color{bleudefrance}254.8 & \color{bleudefrance}0.632 & \color{bleudefrance}0.501 & \color{bleudefrance}1.614 & \color{bleudefrance}276.2 & \color{bleudefrance}0.624 & \color{bleudefrance}0.585 & \color{bleudefrance}2.119\\
    MUNIT' & 361.5 & 0.699 & 0.607 & 1.867 & 289.0 & 0.767 & 0.789 & 2.228\\
     \color{bleudefrance}MUNIT' + \methodall{} & \color{bleudefrance}\textbf{80.4} & \color{bleudefrance}\textbf{0.615}  & \color{bleudefrance}\textbf{0.406}& \color{bleudefrance}\textbf{1.610} & \color{bleudefrance}\textbf{90.8} & \color{bleudefrance}\textbf{0.477} & \color{bleudefrance}\textbf{0.428} & \color{bleudefrance}\textbf{1.689}\\
    
    \bottomrule
  \end{tabular}
  }
  \caption{\methodall{} can improve the performance of MUNIT~\citep{huang2018multimodal}, while for DRIT~\citep{drit} the improvement is marginal. MUNIT' is MUNIT with one more Conv3x3-LN-ReLU layer before the output layer in the decoder, which introduces $0.2\%$ parameters into the generator. Note: for all scores, the lower the better. 
  } 
  \label{tab:drit_munit}
   \vspace{-0.35in}
\end{table}

Interestingly MUNIT, while being larger than DRIT (30M parameters vs. 10M parameters), doesn't perform better on these two datasets. 
One reason for its relatively poor performance could be that the model was not designed for these  datasets (MUNIT uses a much larger unpublished \textit{dogs to big cats} dataset), the dataset are very small, and the default image resolution is slightly different. 
To further improve MUNIT + \methodall{}, we add one more Conv3x3-LN-ReLU layer before the output layer.
Without this, there is only one layer between the outputs and the last re-introduced $\mu$ and $\sigma$.
Therefore, adding one additional layer allows the model to learn a nonlinear function of these $\mu$ and $\sigma$.
We call this model MUNIT' + \methodall{}.
Adding this additional layer significantly enhances the performance of MUNIT while introducing only 75K parameters (about 0.2\%).
We also provide the numbers of MUNIT' (MUNIT with one additional layer) as a baseline for a fair comparison.

Admittedly, the state-of-the-art generative models employ complex architecture and a variety of loss functions; therefore, unveiling the full potential of \methodall{} on these models can be nontrivial and required further explorations. It is fair to admit that the results of all model variations are still largely  unsatisfactory and the image translation task remains an open research problem. 

However, we hope that our experiments on DRIT and MUNIT may shed some light on the potential value of \methodall{}, which could open new interesting directions  of research for neural architecture design.

\section{Conclusion and Future Work}

In this paper, we propose a novel normalization technique, \methodfull{} (\method{}), in combination with a purposely limited variant of shortcut connections, \methodmsfull{} (\methodms{}). When applied to various generative models, we observe that the resulting model is able to preserve  structural aspects of the input, improving the plausibility performance according to  established metrics. 
\method{} and \methodms{} can be implemented in a few lines of code (see Appendix). 
Similar to Instance Normalization, which has been observed to capture the style of image~\citep{huang2017arbitrary,karras2018style,ulyanov2016instance}, \methodfull{} captures structural information. As future work we plan to further explore such disentangling of structural and style information in the design of modern neural architectures. 

It is possible that \method{} and \methodms{} can be applied to a variety of tasks such as image segmentation~\citep{long2015fully,ronneberger2015u}, denoising~\citep{xie2012image,li2017aod}, inpainting~\citep{yu2018generative}, super-resolution~\citep{dong2014learning}, and structured output prediction~\citep{sohn2015learning}.
Further, beyond single image data, \method{} and \methodms{} may also be applied to video data~\citep{NonLocal2018,li2018end}, 3D voxel grids~\citep{DBLP:conf/iccv/TranBFTP15,DBLP:conf/cvpr/CarreiraZ17}, or tasks in natural language processing~\citep{devlin2018bert}.

\subsection*{Acknowledgments}
This research is supported in part by the grants from Facebook, the National Science Foundation (III-1618134, III-1526012, IIS1149882, IIS-1724282, and TRIPODS-1740822), the Office of Naval Research DOD (N00014-17-1-2175), Bill and Melinda Gates Foundation. We are thankful for generous support by Zillow and SAP America Inc.

{\small
\bibliography{reference}
\bibliographystyle{plain}
}

\newpage
\appendix
\appendixpage
\section{Algorithm of \methodall}
The implementation of PONO-MS in TensorFlow~\citep{45381} an PyTorch\citep{paszke2017automatic} are shown in Listing 1 and 2 respectively. 

\definecolor{backcolour}{RGB}{250,250,250}
\definecolor{keywords}{RGB}{255,0,90}
\definecolor{comments}{RGB}{0,0,113}
\definecolor{red}{RGB}{160,0,0}
\definecolor{green}{RGB}{0,150,0}
 
\lstset{language=Python, 
        basicstyle=\ttfamily\small, 
        backgroundcolor=\color{backcolour}, 
        keywordstyle=\color{keywords},
        commentstyle=\color{comments},
        stringstyle=\color{red},
        showstringspaces=false,
        identifierstyle=\color{green},              
        captionpos=b,
        procnamekeys={def,class}
        }
\begin{lstlisting}[language=Python, caption=\method{} and \methodms{} in TensorFlow]
# x is the features of shape [B, H, W, C]

# In the Encoder
def PONO(x, epsilon=1e-5):
    mean, var = tf.nn.moments(x, [3], keep_dims=True) 
    std = tf.sqrt(var + epsilon)
    output = (x - mean) / std
    return output, mean, std
    
# In the Decoder
# one can call MS(x, mean, std)
# with the mean and std are from a PONO in the encoder
def MS(x, beta, gamma):
    return x * gamma + beta
\end{lstlisting}
\label{lst:tf}

\begin{lstlisting}[language=Python, caption=\method{} and \methodms{} in PyTorch]
# x is the features of shape [B, C, H, W]

# In the Encoder
def PONO(x, epsilon=1e-5):
    mean = x.mean(dim=1, keepdim=True)
    std = x.var(dim=1, keepdim=True).add(epsilon).sqrt()
    output = (x - mean) / std
    return output, mean, std
    
# In the Decoder
# one can call MS(x, mean, std)
# with the mean and std are from a PONO in the encoder
def MS(x, beta, gamma):
    return x * gamma + beta
\end{lstlisting}
\label{lst:pytorch}

\section{Equations of Existing Normalization}
Batch Normalization (BN) computes the mean and std across B, H, and H dimensions, i.e.
\[
\mu_c = \mathbb{E}_{b, h, w} [X_{b, c, h, w}], \quad \sigma_c = \sqrt{\mathbb{E}_{b, h, w}[X_{b, c, h, w}^2 - \mu_c] + \epsilon},
\]
where $\epsilon$ is a small constant applied to handle numerical issues.

Synchronized Batch Normalization views features of mini-batches across multiple GPUs as a single mini-batch.

Instance Normalization (IN) treats each instance in a mini-batch independently and computes the statistics across only spatial dimensions, i.e.
\[
\mu_{b, c} = \mathbb{E}_{h, w} [X_{b, c, h, w}], \quad \sigma_{b, c} = \sqrt{\mathbb{E}_{h, w}[X_{b, c, h, w}^2 - \mu_{b,c}] + \epsilon}.
\]

Layer Normalization (LN) normalizes all features of an instance within a layer jointly, i.e.
\[
\mu_{b} = \mathbb{E}_{c, h, w} [X_{b, c, h, w}], \quad \sigma_b = \sqrt{\mathbb{E}_{c, h, w}[X_{b, c, h, w}^2 - \mu_{b}] + \epsilon}.
\]

Finally, Group Normalization (GN) lies between IN and LN, it devides the channels into $G$ groups and apply layer normalization within a group. When $G=1$, GN becomes LN. Conversely, when the $G=C$, it is identical to IN. To define it formally, it computes
\[
\mu_{b, g} = \mathbb{E}_{c \in S_g, h, w} [X_{b, c, h, w}], \quad \sigma_{b, g} = \sqrt{ \mathbb{E}_{c \in S_g, h, w}[X_{b, c, h, w}^2 - \mu_{b,g}] + \epsilon},
\]
where $S_g = \{\lceil\frac{(g-1)C}{G} + 1\rceil , \dots, \lceil \frac{gC}{G} \rceil\}$.

\section{\method{} Statistics of Models Pretrained on ImageNet}
\begin{figure*}
    \centering
    \includegraphics[width=\linewidth]{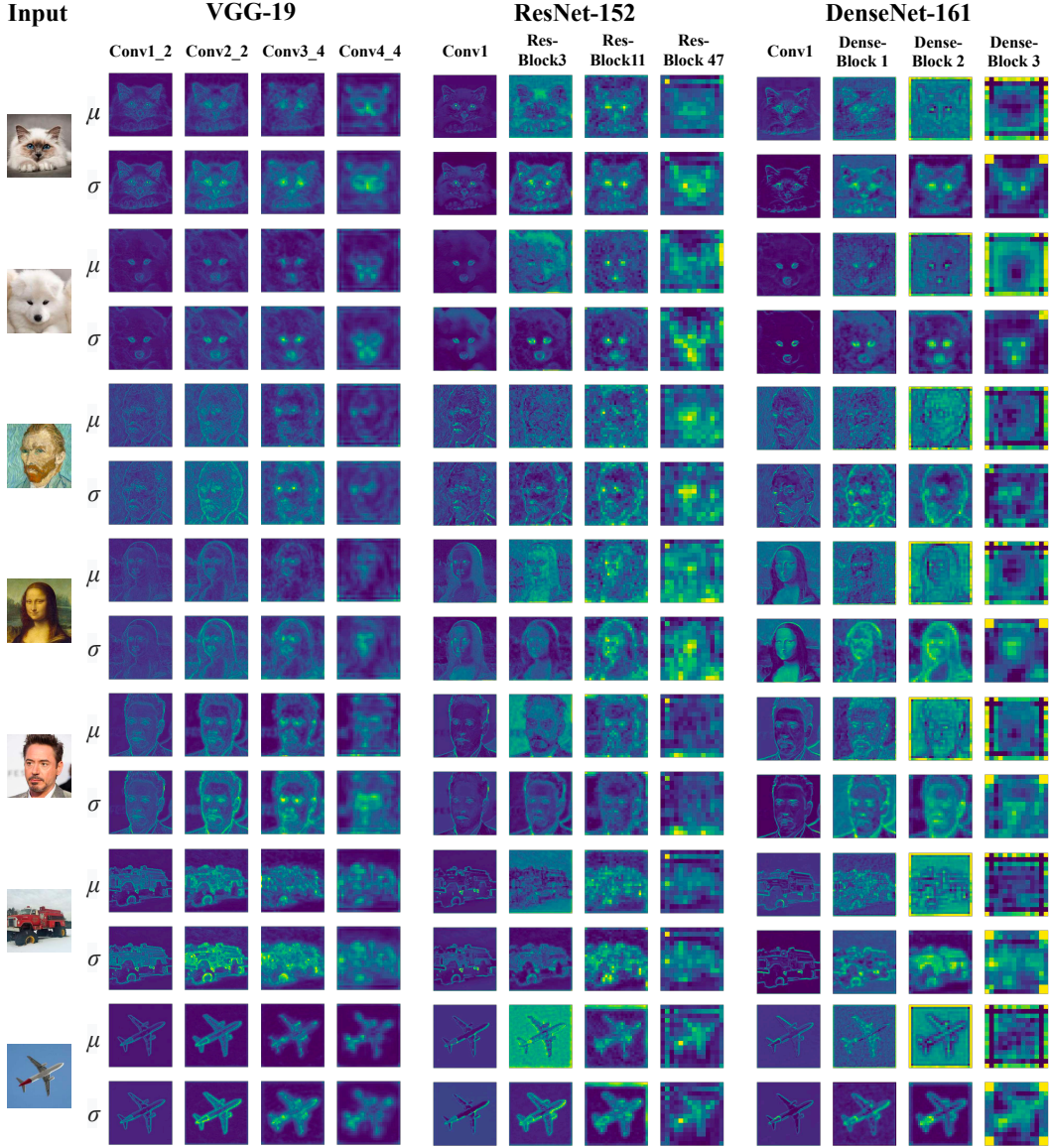}
    \caption{We extract the \method{} statistics from VGG-19, ResNet-152, and Dense-161 at layers right before downsampling (max-pooling or strided convolution).}
    \label{fig:struct_stats}
\end{figure*}

\autoref{fig:struct_stats} shows the means and the standard deviations extracted by \method{} based on the features generated by VGG-19~\citep{simonyan2014very}, ResNet-152~\citep{he2016deep}, and DenseNet-161~\citep{huang2017densely} pretrained on ImageNet~\citep{deng2009imagenet}.

\section{Implementation details}
\begin{figure*}
    \centering
    \includegraphics[width=0.4\linewidth]{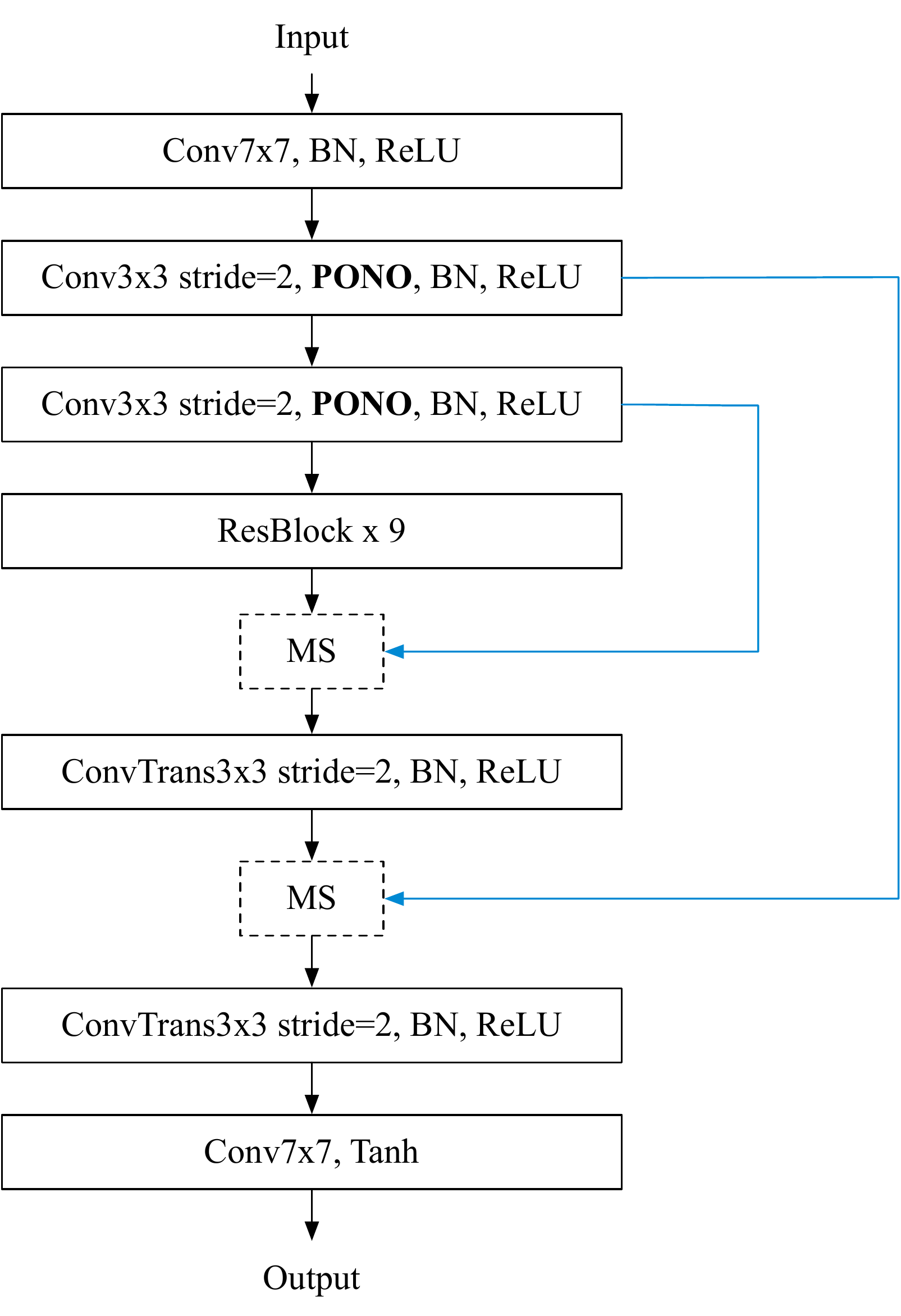}
    \caption{The generator of CycleGAN + \methodall{}. Pix2pix uses the same architecture. The operations in a block is applied from left to right sequentially. The \textbf{\textcolor[HTML]{3498DB}{blue}} lines show how the first two moments are passed. ConvTrans stands for transposed convolution. Each ResBlock has Conv3x3, BN, ReLU, Conv3x3, and BN.}
    \label{fig:cyclegan_arch}
\end{figure*}

We add \method{} to the encoder right after a convolution operation and before other normalization or nonlinear activation function.
\autoref{fig:cyclegan_arch} shows the model architecture of CycleGAN~\citep{zhu2017unpaired} with \methodfull{}. Pix2pix~\citep{isola2017image} uses the same architecture.

\section{Qualitative Results Based on CycleGAN and Pix2pix}
\begin{figure*}
    \centering
    \includegraphics[width=\linewidth]{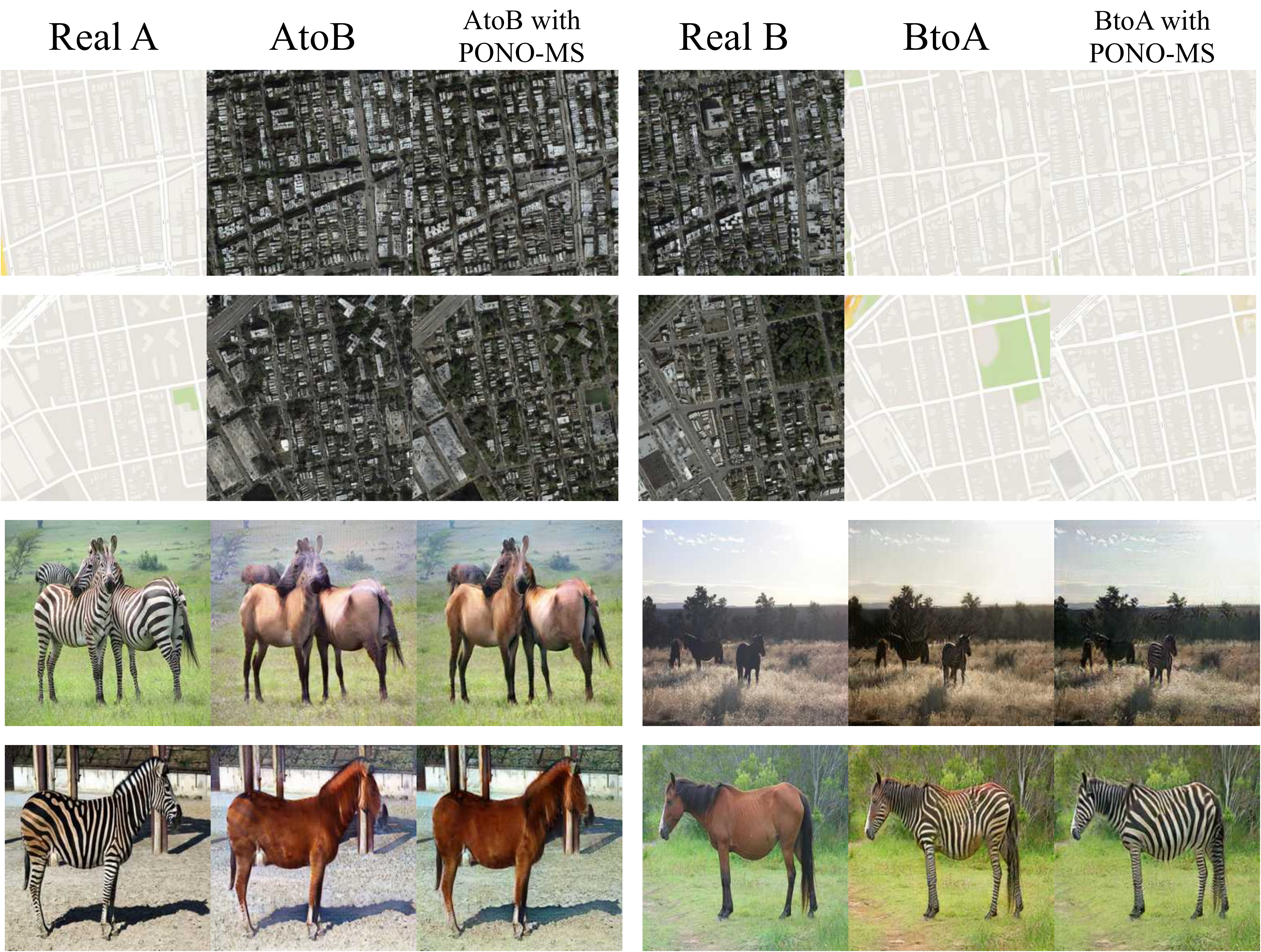}
    \caption{Qualitative results of CycleGAN (with/without PONO-MS) with randomly sampled inputs.}
    \label{fig:appendix_show1}
\end{figure*}

\begin{figure*}
    \centering
    \includegraphics[width=\linewidth]{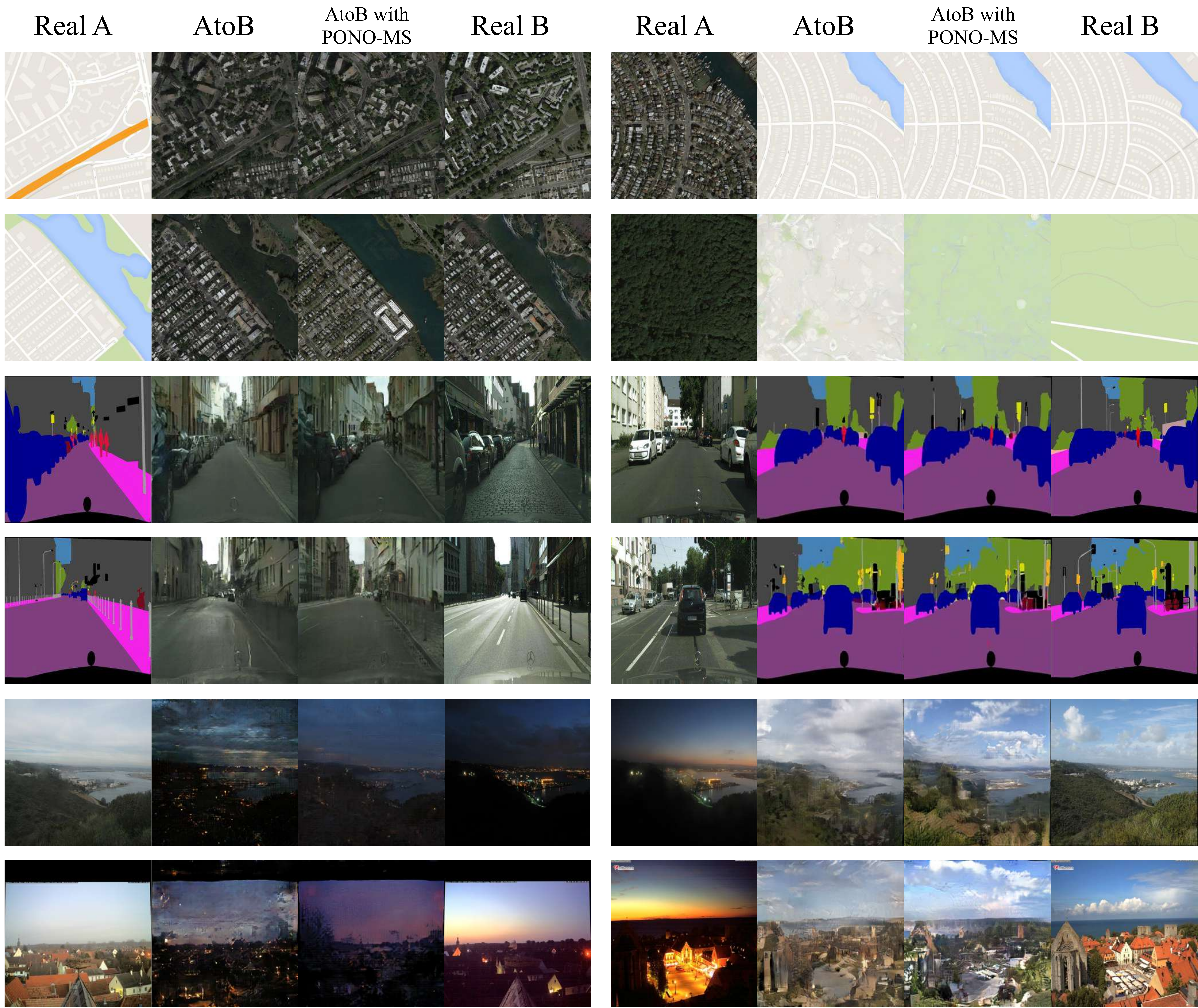}
    \caption{Qualitative results of Pix2pix (with/without PONO-MS) with randomly sampled inputs.}
    \label{fig:appendix_show2}
\end{figure*}

We show some outputs of CycleGAN in \autoref{fig:appendix_show1}.
The Pix2pix outputs are shown in \autoref{fig:appendix_show2}.

\section{Qualitative Results Based on DRIT and MUNIT.}
\begin{figure*}
    \centering
    \includegraphics[width=\linewidth]{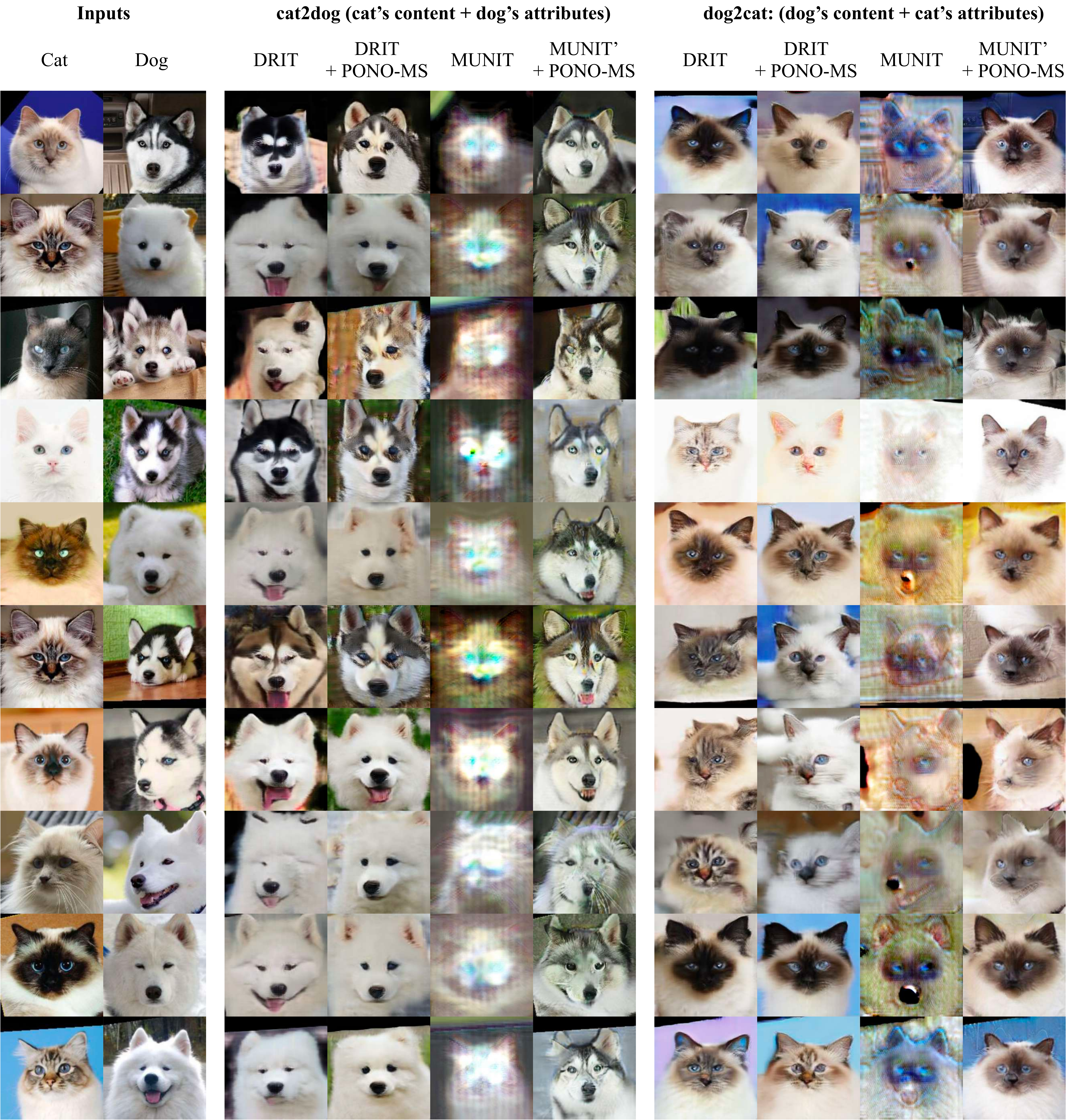}
    \caption{Qualitative results of DRIT and MUNIT (with/without PONO-MS) with randomly sampled inputs.}
    \label{fig:both_example_full}
\end{figure*}

We randomly sample 10 \textit{cat and dog} image pairs and show the outputs of DRIT, DRIT + \methodall{}, MUNIT, and MUNIT' \methodall{} in \autoref{fig:both_example_full}.

\section{\method{} in Image Classification}
\begin{figure*}[t!]
    \centering
    \includegraphics[keepaspectratio,width=5.6in]{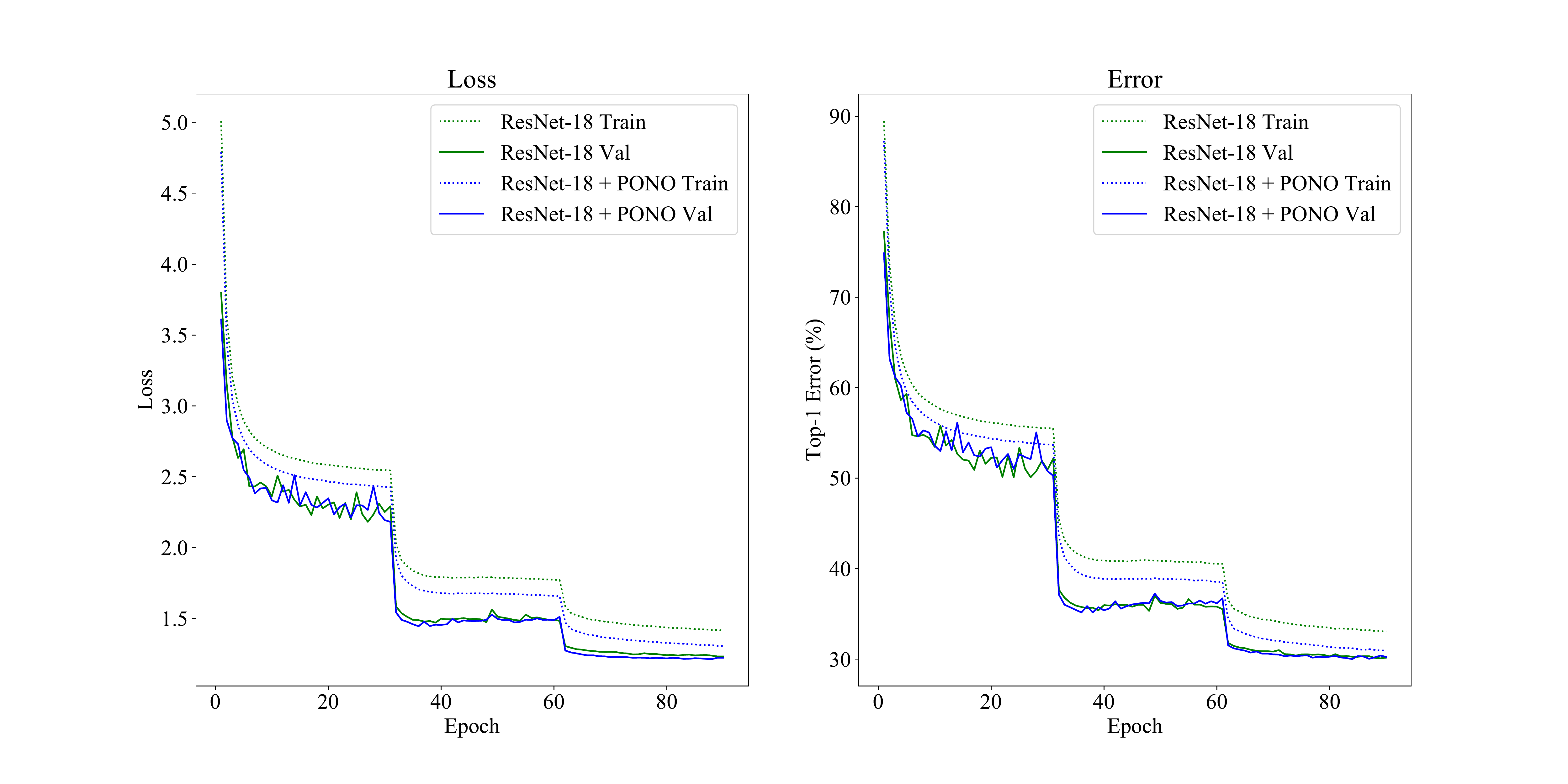}
    \caption{Training and validation curves of ResNet-18 and ResNet-18 + \method{} on ImageNet.}
    \label{fig:resnet18}
\end{figure*}

To evaluate \method{} on image classification task, we add \method{} to the begining of each ResBlock of ResNet-18~\citep{he2016deep} (also affects the shortcut). We followed the common training procedure base on Wei Yang's open sourced code~\footnote{\href{https://github.com/bearpaw/pytorch-classification}{https://github.com/bearpaw/pytorch-classification}} on ImageNet~\citep{krizhevsky2012imagenet}.
\autoref{fig:resnet18} shows that with \method{}, the training loss and error are reduced significantly and the validation error also drops slightly from 30.09 to 30.01.
Admittedly, this is not a significant improvement. We believe that this result may inspire some future architecture design.

\end{document}